\documentclass[lettersize,journal]{IEEEtran}
\usepackage{amsmath,amsfonts}
\usepackage{algorithmic}
\usepackage{algorithm}
\usepackage{array}
\usepackage[caption=false,font=normalsize,labelfont=sf,textfont=sf]{subfig}
\usepackage{textcomp}
\usepackage{stfloats}
\usepackage{url}
\usepackage{verbatim}
\usepackage{graphicx}
\usepackage{cite}

\usepackage{booktabs,chemformula}
\usepackage{multirow}
\usepackage{setspace}
\usepackage{amssymb}	
\usepackage{color, colortbl, soul}
\definecolor{Gray}{gray}{0.95}
\definecolor{LightCyan}{rgb}{0.88,1,1}
\usepackage{arydshln}

\usepackage{pifont}
\usepackage{makecell}

\usepackage{soul}

\newcommand{\etal}{\textit{et al}.}
\newcommand{\ie}{\textit{i}.\textit{e}.}
\newcommand{\eg}{\textit{e}.\textit{g}.}

\begin{document}

\title{AKVSR: Audio Knowledge Empowered Visual Speech Recognition by Compressing Audio Knowledge of a Pretrained Model}
\author{Jeong Hun Yeo, Minsu Kim, Jeongsoo Choi, Dae Hoe Kim, and Yong Man Ro,~\IEEEmembership{Senior Member,~IEEE}

\thanks{J. H. Yeo, M. Kim, J. Choi, and Y. M. Ro are with the Image and Video Systems Laboratory, School of Electrical Engineering, Korea Advanced Institue of Science and Technology (KAIST), Republic of Korea (e-mail: sedne246@kaist.ac.kr; ms.k@kaist.ac.kr; jeongsoo.choi@kaist.ac.kr; ymro@kaist.ac.kr)}
\thanks{D. H. Kim is with the Visual Intelligence Research Section, Superintelligence Creative Research Laboratory, Electronics and Telecommunications Research Institute (ETRI), Republic of Korea (e-mail: dhkim19@etri.re.kr). Corresponding author: Y. M. Ro (fax: 82-42-350-5494)}
}


\maketitle

\begin{abstract}
Visual Speech Recognition (VSR) is the task of predicting spoken words from silent lip movements. VSR is regarded as a challenging task because of the insufficient information on lip movements. In this paper, we propose an Audio Knowledge empowered Visual Speech Recognition framework (AKVSR) to complement the insufficient speech information of visual modality by using audio modality. Different from the previous methods, the proposed AKVSR 1) utilizes rich audio knowledge encoded by a large-scale pretrained audio model, 2) saves the linguistic information of audio knowledge in compact audio memory by discarding the non-linguistic information from the audio through quantization, and 3) includes Audio Bridging Module which can find the best-matched audio features from the compact audio memory, which makes our training possible without audio inputs, once after the compact audio memory is composed. We validate the effectiveness of the proposed method through extensive experiments, and achieve new state-of-the-art performances on the widely-used LRS3 dataset.
\end{abstract}


\section{Introduction}
\IEEEPARstart{V}{isual} Speech Recognition (VSR) is a task of predicting speech content from lip movement without sound. VSR has received a lot of attention due to its practical applications. It can be used as a subtitling tool for silent movies, an auxiliary tool for speech recognition in noisy environments, and a conversation tool for the hearing impaired. 

VSR has significantly improved in its performance along with the development of Deep Learning \cite{ji20123d, vaswani2017attention, cho2014seq2seq, ivanko2022visual, sheng2022importance, sheng2021adaptive, saitoh2017concatenated, sadeghi2020audio,akbari2018lip2audspec}. Many efforts have been made to improve the network architecture of the VSR systems. A visual encoder based on the combination of a 3D convolution layer and a 2D Convolutional Neural Network (CNN) is suggested by \cite{stafylakis2017combining} to encode spatio-temporal visual features from lip movements. To capture the context information from the encoded visual features, prior works \cite{stafylakis2017combining, petridis2018end, zhao2020mi} adopted Recurrent Neural Network (RNN) \cite{mikolov2010recurrent} after the visual encoder. Recently, inspired by the success of the Transformer \cite{vaswani2017attention} in Natural Language Processing (NLP), the VSR model augmented with the Transformer achieved significant speech recognition performances \cite{afouras2018deep, ma2021conformer, ma2022visual, shi2022learning}. Unlike the RNN, the self-attention mechanism of the Transformer enables it to capture dependencies between any two positions in the lip sequence, facilitating a more comprehensive understanding of linguistic content from lip movement. Despite the development of VSR architectures, VSR is still regarded as a challenging task due to the characteristics of visual speech. Different from audio speech, visual speech inherently contains insufficient information to fully represent speech content, as speech is not only produced with the parts that are visible (\ie, lips) but also with diverse internal human organs \cite{sataloff1992human}. Hence, another research stream focuses on complementing insufficient visual information by augmenting the VSR model with additional information.

To complement the insufficient visual information, several prior works proposed to provide audio knowledge into the VSR model. Knowledge Distillation (KD) \cite{hinton2015distilling} is one of the most popular schemes for transferring superior knowledge of a teacher model to a student model. \cite{zhao2020hearing, ren2021learningfromthemaster, afouras2020asrisall, mabrouk2022lip, zhang2023self, elashmawy2021spatio} tried to transfer the audio knowledge of the teacher model into the visual student model. These approaches supervised the student model to follow the soft-label or audio features generated from the teacher model. However, because of the differences in inherent properties between audio and visual modalities called heterogeneity gap \cite{huang2018deep, peng2019cm, lin2020learning}, some knowledge can be discarded during knowledge distillation \cite{ren2021learningfromthemaster}. To bypass the heterogeneity gap, \cite{kim2021cromm, kim2022mvm, yeo2023multi} proposed audio-visual multimodal bridging frameworks based on a memory network \cite{weston2014memory}. They built a visual-to-audio mapping function using a visual key memory and an audio value memory. Through the learned mapping function, the VSR model can utilize the saved audio knowledge. All of the aforementioned methods showed that the VSR systems can better model speech by complementing visual information with audio knowledge. Nevertheless these successes, the previous methods utilizing audio \cite{zhao2020hearing, ren2021learningfromthemaster, afouras2020asrisall, mabrouk2022lip, zhang2023self, elashmawy2021spatio, kim2021cromm, kim2022mvm, yeo2023multi}  do not focus only on transferring linguistic information of audio. For example, prior works \cite{zhao2020hearing, ren2021learningfromthemaster, afouras2020asrisall, mabrouk2022lip, zhang2023self, elashmawy2021spatio} utilized Knowledge Distillation (KD) to make the visual feature to be close to the audio feature without considering the characteristics of audio. An aim of the other works based on memory \cite{kim2021cromm, kim2022mvm, yeo2023multi} also save audio features and reconstruct audio features using visual features without focusing on linguistic information. However, the audio contains not only linguistic information but also contains diverse information such as speaker characteristics, background noises, etc. If we do not consider these diverse factors of audio when using it for VSR training, the complementary effects of audio can be degraded.

Recently, Large-scale pretrained models from raw audio data such as wav2vec2.0 \cite{baevski2020wav2vec} and Hidden-Unit BERT (HuBERT) \cite{hsu2021hubert} achieved significant performance improvement in Audio-based Automatic Speech Recognition (ASR) \cite{lee2008robust, dupont2000audio, pelaez2001recognizing, chibelushi2002review, zheng2021wav}. Especially, HuBERT which is pretrained by masked prediction like BERT \cite{devlin2018bert,zhao2021self} to capture context information from unmasked audio features achieved state-of-the-art performance in ASR after finetuned on paired audio-text dataset. Motivated by the recent success of 1) complementing visual modality with audio modality through audio memory in VSR and 2) self-supervised pretraining in ASR, we try to empower VSR models with audio knowledge extracted from a pretrained model.

In this paper, we propose a novel Audio Knowledge empowered Visual Speech Recognition framework (AKVSR), where the audio knowledge of a large-scale audio pretrained model is extracted with compact representation discarding non-linguistic factors like speaker and noise, and utilized to empower the VSR model. Different from previous approaches complementing visual modality with audio modality \cite{zhao2020hearing, ren2021learningfromthemaster, afouras2020asrisall, mabrouk2022lip, zhang2023self, elashmawy2021spatio, kim2021cromm, kim2022mvm, yeo2023multi}, the proposed method is the first work to adopt the large-scale pretrained audio model in VSR and transfer the linguistic information of audio by considering the properties of audio modality. We would like to highlight that directly utilizing the large-scale pretrained audio model for VSR without considering the characteristics lying in audio such as speaker, and noise may reduce the beneficial effects of using the large-scale pretrained model in VSR. Therefore, we complement the insufficient visual information by using only linguistic information of audio except for other characteristics of audio (i.e., speaker characteristics and noise). To achieve this, the audio knowledge of pretrained HuBERT is vector quantized \cite{van2017neural,esser2021taming, lakhotia2021generative} with a fixed size of clusters which are learned to contain only linguistic information through ASR.R. Therefore, the most representative knowledge in predicting speech can be extracted. The extracted audio knowledge of the pretrained audio model composes the knowledge in compact audio memory, which will be incorporated in VSR models. 
To employ the audio knowledge in training VSR models without input audio, we build an Audio Bridging Module (ABM). ABM is for finding the best-matched audio knowledge with input visual representation from the memory through cross-modal attention. Then, the best-matched audio knowledge is added with encoded visual features to empower speech representations.

Our proposed method has three differences compared to the existing methods based on memory networks \cite{kim2021cromm, kim2022mvm, yeo2023multi}. 1) We utilize rich audio knowledge encoded by a large-scale pretrained audio model and transform the audio knowledge into a compact representation to store only the linguistic information (e.g., phoneme content). 2) The proposed method does not require additional audio input and audio model to supplement insufficient visual representation due to the existence of the ABM when we train the VSR model in contrast to the existing VSR methods. 3) The compact audio memory can be utilized to furnish audio information to any VSR model as the representative knowledge needed for speech prediction, such as phoneme information, does not vary between datasets in speech recognition tasks.

In summary, our key contributions are as follows:
\begin{itemize}
    \item We introduce a novel Audio Knowledge empowered Visual Speech Recognition (AKVSR) framework. To the best of our knowledge, this is the first work to refine non-linguistic factors of audio and transfer the knowledge of a large-scale pretrained audio model into the VSR model.
    
    \item We do not need an additional audio model when we train the VSR model utilizing audio knowledge. 
    
    \item We validate through ASR that the representative knowledge of predicting speech is stored in compact audio memory. Moreover, we verify that the compact audio memory can be adapted to any VSR model.     
    
    \item The proposed AKVSR outperforms the current state-of-the-art VSR model on the most popular sentence-level LRS3 dataset.    
\end{itemize}

\section{Related Works}
\subsection{Visual Speech Recognition}
With the great development of deep learning, many research contributions have been made to VSR, especially in terms of architecture and data.
Chung \etal \cite{chung2016lrw} proposed an English word-level VSR data, LRW, and proposed a VGG-based VSR model. Stafylakis \etal \cite{stafylakis2017combining} improved the architecture of the VSR model by using ResNet-34 \cite{he2016resnet} with one 3D convolution layer and Bi-LSTM. Some works \cite{petridis2018end, tao2020end} proposed an end-to-end Audio-Visual Speech Recognition (AVSR) model, and Petridis \etal \cite{petridis2018end} set a strong baseline in word-level VSR. Some works \cite{weng2019twostream,xiao2020deformation} tried to capture the lip movements in detail by using two-stream networks which utilize both RGB frames and optical flows. Zhao \etal \cite{zhao2020mi} introduced mutual information maximization-based method to enhance the relations of the features with the speech content. Zhang \etal \cite{zhang2020facecutout} proved that using face region instead of using lip region only, is beneficial to VSR. Martinez \etal \cite{martinez2020mstcn} improved the temporal encoding of the back-end by proposing Multi-Scale Temporal Convolutional Network (MS-TCN). Ma \etal \cite{ma2021towards} proposed a distillation-based method of \cite{furlanello2018born} in VSR. They repeatedly trained new models through born-again distillation, where the trained model becomes the new teacher. With the distillation, the VSR model can be lightened without loss of performance. Kim \etal \cite{kim2022speaker} explored speaker dependency of pretrained VSR models and proposed a speaker adaptation method. For the sentence-level VSR, Assael \etal \cite{assael2016lipnet} proposed an end-to-end VSR framework using Connectionist Temporal Classification (CTC) \cite{graves2006ctc}. Chung \etal \cite{chung2017lrs2} improved it to unconstrained sentence-level VSR by proposing LRS2 dataset and sequence-to-sequence architecture \cite{cho2014seq2seq}. Recently, Transformer-based \cite{vaswani2017attention} architectures became the basics for visual speech modeling as they achieved significant VSR performances \cite{afouras2018deep,ma2021conformer,hong2022visual,ma2022visual,koumparoulis2022accurate}. The transformer-based encoder-decoder structure \cite{vaswani2017attention} that utilizes attention mechanisms enables the handling of input and output sequences with varying lengths. This adaptable characteristic empowers the model to accurately transcribe and generate variable-length outputs in the domain of Visual Speech Recognition (VSR).  The proposed method also exploits this useful characteristic to effectively predict different output lengths regardless of the input lengths.

In this paper, we try to improve VSR systems by complementing the limited information of lip movements by proposing a compact audio memory, instead of improving the network architecture. In the next section, we will delve into the recent advancements in incorporating audio information to enhance visual information in the field of VSR.

\subsection{Complementing Visual using Audio in VSR}
There are other efforts trying to augment the VSR model with audio modal knowledge. Afouras \etal \cite{afouras2020asrisall} proposed a method of utilizing a large number of unlabeled audio data. By distilling the predicted logits of a pretrained ASR model into the VSR model, the VSR model can be learned from large-scale unlabelled audio-visual data. On a similar line, \cite{zhao2020hearing,ren2021learningfromthemaster} proposed knowledge distillation methods \cite{hinton2015distilling} by using pretrained ASR models from large-scale audio corpus datasets. By guiding the VSR model to follow the audio features at different levels encoded from the ASR model, the VSR model is expected to extract more discriminative visual features. Another research stream is utilizing memory network \cite{weston2014memory,lee2021videopredictionmem} for saving audio knowledge, Kim \etal \cite{kim2021cromm,kim2021multi} proposed Visual-Audio Memory which can save the audio features during training and read the saved audio knowledge from the learned memory with just visual inputs during inference. They improved the memory network to be able to consider the one-to-many mapping of viseme-to-phoneme by proposing multi-head memory architectures \cite{kim2022mvm}.

Existing methods for VSR utilize the audio modality to complement the visual modality. In this perspective, the KD \cite{afouras2020asrisall, zhao2020hearing, ren2021learningfromthemaster} and memory-based \cite{kim2021cromm, kim2021multi, yeo2023multi} approaches improve the VSR system through multi-modal learning. However, the audio modality contains many characteristics such as speaker, noise, and linguistic content. Transferring audio knowledge without considering non-linguistic factors such as speaker and noise may reduce the complementing effect of the VSR system. Different from the previous approaches, this is the first work to utilize a large-scale pretrained audio model and transfer the audio knowledge to a VSR system considering the linguistic factor of audio modality. Namely, we aim to transfer the audio knowledge focused on only the linguistic factor.

\subsection{Pretraining on Large-scale Databases}
Pretraining neural networks (\eg, BERT \cite{devlin2018bert}) on large-scale datasets achieved significant performances when the pretrained model is adapted to the downstream tasks, in diverse research areas \cite{li2019vlp1,chen2020vlp2,zhou2020vlp3,zhang2020vlp4,dou2022vlp5,sun2019vlp6,radford2021vlp7, zhao2022improving}.

It has also achieved promising results in Automatic Speech Recognition (ASR). Prior works, wav2vec2.0 \cite{baevski2020wav2vec} and HuBERT \cite{hsu2021hubert}, proposed to learn the speech representation from raw audio in a self-supervised manner. Since the methods do not need text annotations, they can be trained via large-scale audio databases. In visual speech modeling, \cite{chung2016syncnet,ma2021lira} proposed self-supervised pretraining methods using audio-visual correspondences. They showed that finetuning the pretrained model on the VSR task can achieve better performance than learning the VSR model from scratch. Recently, AV-HuBERT \cite{shi2022learning} which proposed to pretrain the model with masked predictions using audio-visual databases achieved state-of-the-art performance and showed the powerful speech representation power of the model. Moreover, Zhang \etal \cite{zhang2023self} produces online target features by self-distillation during masked prediction training and then reduces the training cost of self-supervised speech representation learning. 

\begin{figure*}[t!]
	\begin{minipage}[b]{1.0\linewidth}
		\centering
		\centerline{\includegraphics[width=17cm]{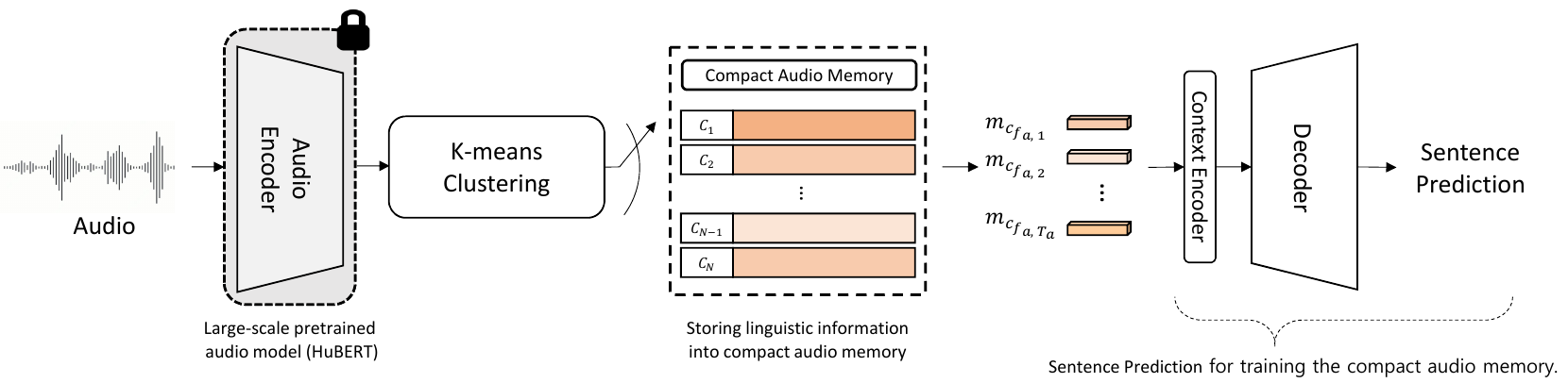}}
	\end{minipage}
	\caption{Overview of building the compact audio memory to store linguistic information of large-scale pretrained audio model. The audio features are generated by the large-scale pretrained audio model, and the features are transformed into discrete representations in compact audio memory. Sentence Prediction is conducted so as to store linguistic information in the compact audio memory. Note that trained compact audio memory is used in the proposed AKVSR.
}
	\label{fig:1}
\end{figure*}


The advantage of utilizing a large-scale pretrained audio model for complementing a visual using audio is that we can acquire improved quality audio knowledge. However, audio knowledge generated by a large-scale pretrained audio model contains a wide range of information, including speech content, speaker characteristics, and noise, as stated in \cite{wang2021fine, lakhotia2021generative}. In this paper, our focus is to store only linguistic audio knowledge in the compact audio memory. Moreover, we inject the audio knowledge from the memory into the visual modality through ABM when we train the VSR model without the audio model.

\section{Methods}

\subsection{Overview} 
The Audio Knowledge empowered Visual Speech Recognition (AKVSR) framework is proposed to improve the complementary effect of audio information in visual speech recognition. Unlike traditional methods, AKVSR obtains improved audio knowledge by adopting a large-scale pretrained audio model and extracts linguistic information by eliminating non-linguistic factors, such as speaker information, through vector quantization. We then store the obtained linguistic information in compact audio memory. This allows us to provide linguistic information of audio knowledge to the VSR model without the additional audio model and inputs. The Audio Bridging Module (ABM) is employed to find the most appropriate audio knowledge from the compact audio memory matched with the visual feature. Therefore, it is possible to complement insufficient visual information by injecting the found linguistic information of the audio knowledge into the VSR model. In the following subsections, we will provide details of each proposed method.

\subsection{Compressing Linguistic Audio knowledge into Compact Audio Memory}
Recent studies \cite{lakhotia2021generative, chang2023exploration, lee2021direct} report that the vector quantized self-supervised speech representation can disentangle linguistic content from speaker characteristics and noise. They show that the same linguistic content can be obtained at least the content is the same, even if the speaker is changed. The rationale behind this lies in the application of vector quantization to speech features at each time step. This process compels the resulting quantized speech feature to exhibit the most discriminative representations of self-supervised speech feature. Here, the most discriminative representation of self-supervised speech feature indicates the linguistic information which is the phonetic information, which is described in \cite{pasad2023comparative}. Motivated by evidence of the prior works, we separately train the K-means clustering and compact audio memory module, and extract the linguistic information from the self-supervised speech representation model through vector quantization.

To this end, as shown in Fig \ref{fig:1}, from a given speech utterance $x_a \in \mathbb{R}^T$, where $T$ is the length of the speech utterance, we encode the audio features $f_a = \{f_{a,i}\}_{i=1}^{T_a} \in \mathbb{R}^{T_a \times d_{a}}$ using a large-scale pretrained audio model $E_a$, where $T_a$ is the frame length of the audio feature, and $d_a$ is the embedding dimension of the audio feature. This process can be expressed as $f_a = E_a(x_a)$. Since we utilize pretrained audio model on large audio data, the extracted features can be regarded as containing rich speech knowledge.

Next, we aim to store linguistic information at the phoneme level by refining the audio features. To achieve this, we first create a k-means clustering model using an audio corpus dataset consisting of a single speaker to minimize non-linguistic factors about speakers inspired by \cite{lakhotia2021generative}. Then, we cluster the audio features $f_a$ into $N$ clusters, $C = [1, 2, ..., N]$, through the k-means clustering model. The cluster labels of each audio feature are determined by the frame-wise quantization, $\textbf{q}(\cdot)$, as follows:
\begin{equation}
    c_{f_{a}}= \textbf{q}(f_{a}) = \{c_{f_{a,i}}\}_{i=1}^{T_a} 
\end{equation}
where $c_{f_{a,i}} \in C$ is the cluster label of the each audio feature.

After the clustering step, we introduce a trainable compact audio memory to store linguistic information (\ie, representative speech feature) for each cluster group. The compact audio memory $M_{a} = \{m_{n}\}_{n=1}^{N}$ is comprised of $N$ discrete representations equal to the number of cluster groups. Each representation denoted $m_n$ has an embedding dimension of $d$. By utilizing the cluster labels generated from the audio features, we are able to access each slot of the compact audio memory. The process of accessing the memory can be generalized as follows:
\begin{equation}
m_{c_{f_{a}}}= \textbf{M}(c_{f_{a}}) = \{m_{c_{f_{a,i}}}\}_{i=1}^{T_a}
\end{equation}
where $\textbf{M}$ is the frame-wise memory accessing function.
For instance, $\textbf{M}(c_{f_{a,i}}\text{ = }1) = m_{1}$ represents that the discrete representation has been extracted from the first slot of the compact audio memory. 

Finally, to store representative knowledge (\ie, linguistic information only) in predicting speech at compact audio memory while discarding non-linguistic information, we perform ASR using the memory through paired audio-text data. To conduct ASR, we employ a context encoder $E_{c}$ and a decoder $D$. When the discrete representations are extracted from the compact audio memory, the context encoder captures the context information between these representations. The contextualized representations are then used to predict the speech content $\hat{y}$ through the decoder as follows: $\hat{y} = D(E_{c}(m_{c_{f_{a}}}))$. To train all components of our model, we use a hybrid CTC/attention loss \cite{watanabe2017hybrid}, which is a commonly used loss function for the speech recognition task. Further details about the losses are provided in subsection \ref{subsection:liptotext}. By performing ASR with the compact audio memory, we can extract and store the linguistic knowledge of pretrained audio model in compact audio memory while disentangling the non-linguistic information which is not important in predicting speech.

\begin{figure*}[t!]
	\begin{minipage}[b]{1.0\linewidth}
		\centering
		\centerline{\includegraphics[width=15cm]{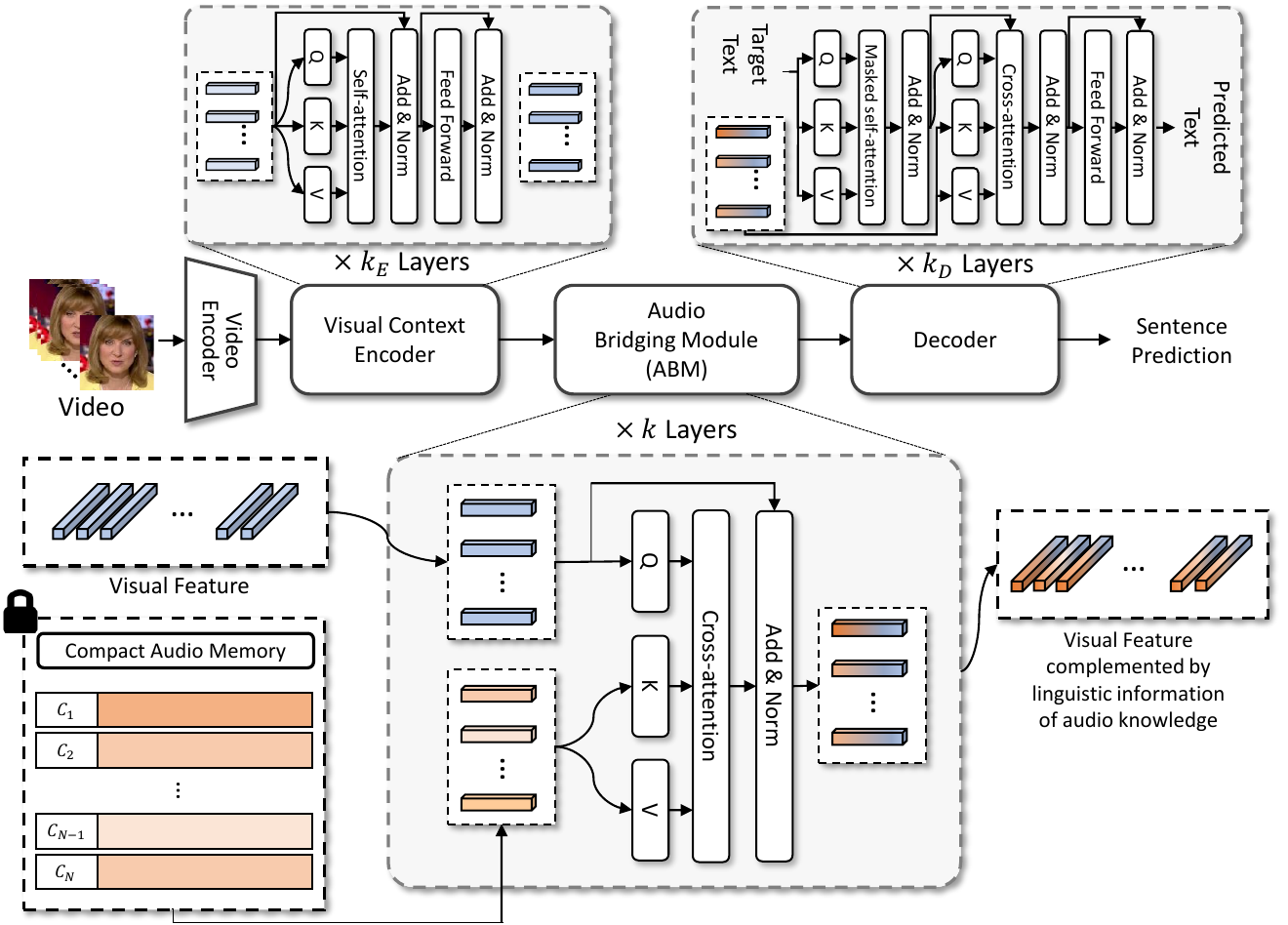}}
	\end{minipage}
	\caption{The overall framework of a proposed AKVSR for complementing visual modality with audio modality. The AKVSR mainly consists of 2 parts: 1) The compact audio memory provides linguistic information from audio knowledge generated by a large-scale pretrained audio model. The meaning of N in compact audio memory is the number of discrete representations in the memory. Moreover, the number of discrete representations is the same as the number of clustering groups of audio features. 2) The proposed ABM finds best-matched information in compact audio memory and injects the linguistic information into the visual feature to complement the insufficient information of lip movements.
}
\vspace{-0.5cm}
	\label{fig:2}
\end{figure*}


\subsection{Injecting Audio Knowledge into VSR with Audio Bridging Module}

Our prior discussion addressed the methods for storing linguistic information within a compact audio memory. This section describes how the saved linguistic audio knowledge can be employed for VSR. To this end, we propose Audio Bridging Module (ABM), which aims to inject the best-matched audio knowledge saved in the memory with visual features into the VSR model to complement insufficient visual modality.

The ABM is trained to identify the most relevant audio knowledge stored in the compact audio memory. Once training is completed, the most suitable audio can be injected into the VSR model through the ABM to enhance the limited lip movement information during inference, even without the presence of audio input data. Therefore, the proposed method uses only visual input for extracting audio knowledge from the compact audio memory and training the VSR model. This is different from the existing VSR methods that require both modal inputs during training for Knowledge Distillation (KD) or contrastive learning. The entire pipeline for finding audio knowledge and complementing the visual information is illustrated in Fig. 2.

Given the lip video $x_{v} \in \mathbb{R} ^{T_{v} \times C \times H \times W}$, where $T_{v}$ is the number of frames of video, $C$ is the channel dimension, $H$ is the height of lip video, and $W$ is the width of lip video. We encode the lip video to visual features through the video encoder and visual context encoder. The video encoder captures compact spatio-temporal information from lip movements \cite{ji20123d}. Next, the visual context encoder can consider the context information of neighboring words via a self-attention mechanism \cite{vaswani2017attention}. The encoding process can be formulated as: $f_{v} = E_{v}(x_{v}) = \{ f_{v,i} \}_{i=1}^{T_{v}}$ where $E_v$ indicate both the video encoder and visual context encoder, $f_{v,i} \in \mathbb{R}^{d}$ is visual feature, and $d$ represents the embedding dimension.

We construct the ABM to find the best-matched audio knowledge with visual features $f_v$ from the compact audio memory utilizing a cross-attention mechanism. The primary motivation behind employing cross-attention in our model is to receive the linguistic information from unaligned compact audio memory for complementing visual features. While compact audio memory is comprised of a fixed number of N discrete representations, the number of visual features varies when the input video changes. Recent study\cite{tsai2019multimodal} shows that cross-attention can be used for aligning multi-modal language sequences, and receiving information from another modality to supplement one modality. Motivated by the success of these works, we utilize attention scores based on cross-attention to complement the visual features through compact audio memory. Moreover, through this approach, the correlation between unaligned visual features and discrete representations of compact audio memory can be calculated. In cross-attention, the attention score measures how much the given visual feature correlates to the discrete representation of compact audio memory.

To calculate the attention scores, we define the visual query of $i$-th visual feature as $Q_{f_{v,i}}=f_{v,i}W_{Q_{v}}$, and audio keys of $j$-th linguistic information in compact audio memory as $K_{a,j}=m_{j}W_{K_{a}}$, where $W_{Q_{v}} \in \mathbb{R}^{d \times d_{k}}$, and $W_{K_{a}} \in \mathbb{R}^{d \times d_{k}}$  are weight matrices. Then, to find appropriate linguistic information through visual features, we calculate the attention score $A_{i,j}^{(0)}$ between a $i$-th visual feature $f_{v, i}$ and each audio feature stored in compact audio memory as follows: 

\begin{equation}
\begin{split}
    A_{i,j}^{(0)} &= Softmax(\frac{Q_{f_{v,i}}K_{a,j}^T}{\tau}) \\
    & = \frac{exp({f_{v,i}W_{Q_{v}}}W_{K_{a}}^{T}m_{j}^{T}/\tau)}{\Sigma_{n=1}^{N}exp({f_{v,i}W_{Q_{v}}}W_{K_{a}}^{T}m_{n}^{T}/\tau)},
\end{split}
\end{equation}

\noindent
where the $\tau$ is a scaling parameter. Intuitively, the attention score provides how much linguistic information in the $j$-th slot in compact audio memory is related to $i$-th visual feature. For example, the higher attention score can load more audio information from one of the N discrete representations in compact audio memory. Moreover, we would like to emphasize that the proposed method uses only visual input to recall the audio information inputs because N discrete units of compact audio memory can represent all audio features.

The linguistic audio knowledge in the compact audio memory for complementing $i$-th visual feature can be reconstructed by using the attention score, which can be denoted as: 
 
\begin{equation}
    m'^{(0)}_{f_{v,i}} = \Sigma_{j=1}^{N} A^{(0)}_{i,j}(m_{j}W_{V_{a}}),
\end{equation}

\noindent
where $W_{V_{a}} \in \mathbb{R}^{d \times d_{v}}$ is a weight matrix. Then, we employ a weight matrix $W_o \in \mathbb{R}^{d_{v} \times d}$ to match the dimensions of the reconstructed audio knowledge and visual feature. Moreover, since one linguistic audio knowledge is reconstructed for each visual feature, an alignment process between visual and audio features is not required. Therefore, we utilize the reconstructed audio knowledge for the $i$-th visual feature as follows:   

\begin{equation}
    f_{v,i}^{(1)} = LN( f_{v,i} + m'^{(0)}_{f_{v,i}}W_{o} ),
\end{equation}

\noindent
where $f_{v,i}^{(1)} \in \mathbb{R}^{d_{v}}$ is the visual feature complemented once by the information of compact audio memory, and the $LN$ denotes the Layer Normalization \cite{ba2016layernorm}.

We can generalize the above processes as consisting of two parts. 1) We bring the linguistic information from compact audio memory via a cross-attention mechanism, and 2) we inject the information into the visual feature sequence.  To this end, we define the visual feature sequence complemented $k-1$ times as $f^{(k-1)}_{v} \in \mathbb{R}^{T_{v} \times d}$ and denote the query of visual feature sequence $Q_{{f}_{v}}^{(k-1)}=f^{(k-1)}_{v}W_{Q_{v}}^{(k-1)}$, the entire audio keys $K_{a}^{(k-1)}=M_{a}W_{K_{a}}^{(k-1)}$, and the audio values of entire compact audio memory as $V_{a}^{(k-1)} = M_{a}W_{V_{a}}^{(k-1)}$. Then, we can formulate the first process of getting linguistic audio knowledge from compact audio memory as follows:  
\begin{equation}
    \begin{split}
         m_{f_{v}}^{\prime (k-1)}  &= A^{(k-1)} V_{a}^{(k-1)} \\  
         &= Softmax(\frac{Q_{f_{v}}^{(k-1)}{K_{a}^{(k-1)}}^{T}}{\tau})V_{a}^{(k-1)} 
    \end{split}
\end{equation}

where $A^{(k-1)} = \{A_{i}^{(k-1)} \}_{i=1}^{T_{v}}$ is attention scores between visual features sequence and every slot in compact audio memory. We then complement the visual feature sequence with knowledge brought by attention scores from compact audio memory as follows: $f_{{v}}^{(k)} = LN( f_{{v}}^{(k-1)} + m_{f_{v}}^{\prime(k-1)}W_{o}^{(k-1)} )$

\subsection{Lip-To-Text Translation}
\label{subsection:liptotext}
In the previous section, the proposed ABM complements the visual feature through the linguistic information stored in compact audio memory. In this section, we introduce the process of lip-to-text translation through the complemented visual feature.

For the decoder, we use transformer, following the previous methods, to predict sentences. Different from previous works, our proposed method can provide additional linguistic audio knowledge to the visual feature sequence. Therefore, the complemented visual feature is fed into the decoder as input, and the decoder predicted $L$ subwords through the features. We then employ the hybrid CTC/attention \cite{watanabe2017hybrid} loss to supervise the proposed model through the predicted sentence and target sentence.

CTC \cite{graves2006ctc} loss is widely used to guide the VSR model through frame-wise prediction based on conditional independence. The frame-wise posterior distribution $p(s_{t}|\textbf{x})$, where the $s_{t}$ is the target subword corresponding to $t$-th frame and $\textbf{x}$ is a visual input. The CTC probability and CTC loss can be formulated as follows:  
\begin{align}
    p_{ctc}(\textbf{s}|\textbf{x}) &\approx \sum_{s}\prod_{t=1}^{T}p(s_{t}|\textbf{x}) \\
    \mathcal{L}_{ctc} &= \log p_{ctc}(\textbf{s}|\textbf{x})
\end{align}
where the $\textbf{s}$ is a target subwords containing blank symbols. Attention loss is employed to learn an implicit language model. The decoder infers the next target subword conditioned on the previous prediction. The attention loss can be formulated as:
\begin{align}
    p_{att}(\textbf{s}|\textbf{x}) &= \prod_{l=1}^{L}p(s_{l}|s_{1}, ..., s_{l-1}, \textbf{x}) \\
    \mathcal{L}_{att} &= \log p_{att}(\textbf{s}|\textbf{x})
\end{align}
 where $s_{l}$ is the predicted subword at time step $l$. Finally, the hybrid CTC/Attention loss can be formulated by weighted summation of the two loss functions:   
\begin{equation}
    \mathcal{L}_{tot} = (1-\lambda)\mathcal{L}_{att} + \lambda\mathcal{L}_{ctc},
\end{equation}
where $\lambda$ is the balancing weight.
\section{Experimental Setup}

\subsection{Dataset.}

\noindent \textbf{LRS2 \& LRS3} are two of the most popular publicly available sentence-level VSR datasets. LRS2 \cite{chung2017lrs2} and LRS3 \cite{afouras2018lrs3} datasets are extracted from BBC television and TED \& TEDx talks, respectively. The LRS2 consists of 28 hours of video for training and 195 hours of video for pretraining. The difference between the training dataset and the pretraining dataset is the duration of each video clip. The duration of pretraining videos is longer than the trainset. In the same way, the LRS3 comprised 30 hours of video for training and 403 hours of video for pretraining. We divide the LRS3 dataset into 30 hours of video for low-resource settings and 433 hours of video for high-resource settings to verify the effectiveness according to the amount the labeled data like AV-HuBERT \cite{shi2022learning}. In addition, The LRS2 dataset is divided into 28 hours of video for low-resource settings and 223 hours of video for high-resource settings. 

We follow the preprocessing pipeline of AV-HuBERT \cite{shi2022learning}. Firstly, we extract the landmarks from each video clip using dlib \cite{king2009dlib}, and employ affine transformation to align each frame to the reference face frame. Next, we crop the video into 96 $\times$ 96 corresponding to the lip region. For data augmentation at train time, the random crop and random horizontal flip are used.

\begin{table*}[h]
  \renewcommand{\arraystretch}{1.1}
  \renewcommand{\tabcolsep}{5mm}
  \centering
  \caption{VSR performance comparison between the proposed methods and previous models on the LRS3 dataset. $\dagger$ indicates that the non-public video-text data is used for training the VSR model.  $\ddagger$ indicates that the SynthVSR uses an additional 3,652 hours of synthetic video-text data to train the Conformer-BASE model. (+$\alpha$) indicates the amount of pseudo-text labels generated by the pretrained ASR model.}
  \resizebox{0.9\linewidth}{!}{
  \begin{tabular}{cccccc}
    \hline \hline
    Method  & Backbone   & \makecell{Labeled \\ word \\ data (hrs)}  & \makecell{Labeled \\ sentence \\ data (hrs)}  & \makecell{Unlabeled \\ sentence \\ data (hrs)}  & WER(\%) \\ \hline
    Afouras et al. \cite{afouras2020asrisall} & CNN  & 157 & 433 & - & 68.8 \\
    Zhang et al. \cite{zhang2019spatio} & CNN  & 157 & 698 & - & 60.1 \\
    Afouras et al. \cite{afouras2018lrs3} & Transformer  & 157 & 1362 & - & 58.9 \\
    Xu et al. \cite{xu2020discriminative} & RNN  & 157 & 433 & - & 57.8 \\
    Shillingford et al. \cite{shillingford2018large} & RNN & - & 3.886 & - & 55.1 \\
    Ma et al. \cite{ma2021conformer} & Conformer  & - & 433 & - & 46.9 \\
    Ma et al. \cite{ma2021conformer} & Conformer  & 157 & 433 & - & 43.3 \\
    Prajwal et al. \cite{prajwal2022sub} & Transformer  & - & 698 & - & 40.6 \\
    Ma et al. \cite{ma2022visual} & Conformer  & - & 433 & - & 37.9 \\
    Ma et al. \cite{ma2022visual} & Conformer  & 157 & 433 & - & 35.1 \\
    Makino et al.$^{\dagger}$ \cite{makino2019recurrent} & RNN & - & 31,000 & - & 33.6 \\
    Serdyuk et al.$^{\dagger}$ \cite{serdyuk2022transformer} & Conformer  & - & 90,000 & - & 17.0 \\
    \hline
        \multirow{2}{*}{SynthVSR \cite{liu2023synthvsr}}  
                            & Conformer-BASE  & - & 30 & 3652$^{\ddagger}$ & 43.3 \\
                            & Conformer-BASE & - & 433 & 3652$^{\ddagger}$ & 27.9 \\
    \hline

    \multirow{5}{*}{AV-HuBERT \cite{shi2022learning}}  & Transformer-BASE  & - & 30 & 1759 & 46.1 \\
                            & Transformer-BASE  & - & 433 & 1759 & 34.8 \\
                            & Transformer-LARGE  & - & 30 & 1759 & 32.5 \\
                            & Transformer-LARGE & - & 433 & 1759 & 28.6 \\
                            & Transformer-LARGE & - & 433(+1,326) & 1759 & 26.8 \\
    \hline
        \multirow{3}{*}{Lohrenz. et al. \cite{lohrenz2023relaxed}}  
                            & Transformer-LARGE  & - & 30 & 1326 & 44.0 \\
                            & Transformer-LARGE & - & 433 & 1326 & 28.8 \\
                            & Transformer-LARGE & - & 433(+1,326) & 1326 & 26.3 \\
                            
    \hline
        \multirow{3}{*}{RAVEn \cite{haliassos2022jointly}} 
                            & Transformer-LARGE  & - & 30 & 1759 & 32.5 \\
                            & Transformer-LARGE & - & 433 & 1759 & 27.8 \\
                            & Transformer-LARGE & - & 433(+1,326) & 1759 & 24.4 \\

    \hline

    \multirow{5}{*}{\textbf{Proposed Method}}  & Transformer-BASE  & - & 30 & 1759 & \textbf{41.6} \\
                            & Transformer-BASE  & - & 433 & 1759 & \textbf{34.2} \\
                            & Transformer-LARGE  & - & 30 & 1759 & \textbf{29.1} \\
                            & Transformer-LARGE  & - & 433 & 1759 & \textbf{27.6} \\ 
                            & Transformer-LARGE  & - & 433(+1,326) & 1759 & \textbf{23.6} \\    
    \hline \hline
  \end{tabular}}
  \label{tab:1}
\end{table*}


\begin{table*}[h]
	\renewcommand{\arraystretch}{1.1}
	\renewcommand{\tabcolsep}{5mm}
  \centering
  \caption{VSR performance comparison between the proposed methods and previous models on LRS2 dataset.}
  \resizebox{0.9\linewidth}{!}{
  \begin{tabular}{cccccc}
    \hline \hline
    Method  & Backbone   & \makecell{Labeled \\ word \\ data (hrs)}  & \makecell{Labeled \\ sentence \\ data (hrs)}  & \makecell{Unlabeled \\ sentence \\ data (hrs)}  & WER(\%) \\ \hline
    Afouras et al. \cite{afouras2020asrisall} & CNN  & 157 & 223 & - & 58.5 \\
    Zhang et al. \cite{zhang2019spatio} & CNN  & 157 & 698 & - & 51.7 \\
    Afouras et al. \cite{afouras2018lrs3} & Transformer  & 157 & 1362 & - & 48.3 \\
    Kim et al. \cite{kim2021cromm} & Transformer  & 157 & 656 & - & 46.2 \\
    Kim et al. \cite{kim2022mvm} & Transformer  & 157 & 656 & - & 44.5 \\
    Ma et al. \cite{ma2021conformer} & Conformer  & - & 223 & - & 39.1 \\
    Ma et al. \cite{ma2021conformer} & Conformer  & 157 & 223 & - & 37.9 \\
    Ma et al. \cite{ma2022visual} & Conformer  & - & 223 & - & 32.9 \\
    K R Prajwal et al. \cite{prajwal2022sub} & Transformer  & - & 698 & - & 28.9 \\
    Ma et al. \cite{ma2022visual} & Conformer  & 157 & 223 & - & 28.7 \\
    \hline
    \multirow{4}{*}{AV-HuBERT \cite{shi2022learning}}  & Transformer-BASE  & - & 28 & 1759 & 43.3 \\
                            & Transformer-BASE  & - & 223 & 1759 & 31.2 \\
                            & Transformer-LARGE  & - & 28 & 1759 & 32.2 \\
                            & Transformer-LARGE & - & 223 & 1759 & 25.5 \\                     
    \hline
    \multirow{4}{*}{\textbf{Proposed Method}}  & Transformer-BASE  & - & 28 & 1759 & \textbf{40.5} \\
                            & Transformer-BASE  & - & 223 & 1759 & \textbf{30.1} \\
                            & Transformer-LARGE  & - & 28 & 1759 & \textbf{28.7} \\
                            & Transformer-LARGE  & - & 223 & 1759 & \textbf{24.1} \\    
    \hline \hline
  \end{tabular}}
  
  \label{tab:2}
\end{table*}


\subsection{Implementation Details.}
For the VSR model, we employ a state-of-the-art architecture based on transformer encoder-decoder models detailed in \cite{shi2022learning}. The video encoder network consists of a 3D CNN, ResNet-18, and the visual context encoder based on the transformer encoder. We employ a Transformer BASE  (12 layers) and a Transformer LARGE (24 layers)  same as AV-HuBERT \cite{shi2022learning}, which models have 103M and 325M of the number of parameters, respectively. We initialize the parameters of both the video encoder and visual context encoder from the BASE model and the LARGE model of AV-HuBERT, respectively. Similar to the transformer encoder, there are two versions of the decoder. transformer decoders have 6 layers for the BASE and 9 layers for the LARGE. We then utilize a unigram-based subword unit \cite{kudo2018subword} composed of 1000 subwords like Av-HuBERT to decode the features produced by the encoder into subword units. We initialize the parameters of the transformer decoders fine-tuned on the audio corpus of LRS2 and LRS3, respectively. The proposed compact audio memory is constructed of 200 discrete units. The embedding dimension of the memory for BASE and LARGE are 768 and 1024, and the number of parameters are 0.15M and 0.2M, respectively. To train the compact audio memory with both context encoder and decoder through the ASR task, we use four transformer layers for context encoder, and transformer decoders (BASE) such as the VSR model decoder. We then use 2 cross-attention layers with 8 multi-head attention mechanisms for ABM. The ABM has 4.7M parameters. The balancing weight $0.1$ is used at the proposed method.

\section{Experimental results}
In this section, we validate the effectiveness of the AKVSR framework utilizing both compact audio memory and ABM through a comparison of the proposed method with the previous VSR methods. Then, to demonstrate the availability of each component, we provide various experimental results. Firstly, we examine how the linguistic information constructed by different pretrained audio model affects the VSR models when we inject the information into the visual modality. Next, we investigate the impact on VSR according to the amount of injected audio information through the proposed ABM.  Finally, we demonstrate that the AKVSR can complement visual information of the other VSR models.  

\subsection{Comparisons with the State-of-the-art}
To verify the effectiveness of the AKVSR, we first compared it to AV-HuBERT on the LRS3 dataset. We adopt the AKVSR to both the Transformer BASE and Transformer LARGE models. To compare performance based on the amount of video-text label data, we perform fine-tuning using both low-resource (30 hours) and high-resource (433 hours) settings. Thus, we conduct experiments with four different settings on the LRS3 dataset. 

Table \ref{tab:1} presents the results of the AVKSR method on the LRS3 dataset. In the \cite{serdyuk2022transformer}, the state-of-the-art model using a conformer encoder as a visual front-end shows a best WER of 17.0\%. However, we would like to emphasize that the state-of-the-art performance model is trained on 90,000 hours of non-public video-text data. On the other hand, we use publicly available video-text data for training. Therefore, we would like to note that a direct comparison between the method \cite{serdyuk2022transformer} (trained on 90,000 hours of data) and the proposed method is not fair in terms of state-of-the-art (SOTA) comparison. Instead, to expand the experimental dataset, we utilize an additional VoxCeleb2 dataset, which datasets comprise 1,326 hours of video. Similar to recent works \mbox{\cite{shi2022learning, haliassos2022jointly}}, we generate pseudo-text labels from the VoxCeleb2 Dataset.

In the LRS3 dataset, the detailed comparison of our proposed methods' performances with recent state-of-the-art performances is shown in Table \ref{tab:1}. We would like to emphasize the proposed methods consistently outperform the AV-HuBERT model across various training data sizes. When trained on a 30-hour dataset, the proposed model achieves a WER of 29.1\%, which is notably lower than the 32.5\% WER of AV-HuBERT. Similarly, with a larger dataset of 433 hours, the proposed method shows an improved WER of 27.6\% compared to AV-HuBERT's 28.6\%. This trend of enhanced performance continues with the most extensive dataset of 433(+1,326) hours, where the proposed method attains a WER of 23.6\%, surpassing AV-HuBERT's 26.8\%. Moreover, we compare this performance of the proposed model with the recent state-of-the-art methods \mbox{\cite{haliassos2022jointly, lohrenz2023relaxed}}. Our methods achieve a WER of 23.6\%, surpassing the recent Raven\cite{haliassos2022jointly} method, which records a WER of 24.4\%. Additionally, it outperforms the previous approach \cite{lohrenz2023relaxed} relaxed attention, which has 26.3\% WER. Please note that these reported WERs of \cite{shi2022learning, liu2023synthvsr, haliassos2022jointly, lohrenz2023relaxed} are based on not using the language model during the inference stage for fair comparison.

Regarding correlated results to the model parameters, the LARGE model surpasses 12.5\% WER compared to the BASE model since the LARGE model has 122M more parameters. On the other hand, the baseline method achieves a WER of 46.1\% and 32.5\%. We would like to emphasize that the compact audio memory and ABM utilize only around 5M additional parameters compared to the baseline.

To validate the effectiveness of the proposed method in another dataset, we additionally conduct experiments on the LRS2 dataset. The AV-HuBERT method does not provide results for the LRS2 dataset, because the video data of LRS2 is open for academic research. In our comparative analysis showcased in Table \mbox{\ref{tab:2}}, the performance of the AV-HuBERT model and our proposed method across different configurations is detailed. For the Transformer-BASE configuration, AV-HuBERT achieves a WER of 43.3\% with 28 hours of training data and 31.2\% with 223 hours, while our proposed method shows improved results with a WER of 40.5\% and 30.1\%. In addition, the proposed methods also outperform the AV-HUBERT by achieving a WER of 24.1\%, in the Transformer-LARGE configuration.

\begin{table}[]
    \renewcommand{\arraystretch}{1.2}
    \renewcommand{\tabcolsep}{9.3mm}
    \centering
    \caption{VSR performance of the proposed AKVSR when different pretrained audio models (CPC, Wav2vec2.0, and HuBERT) are utilized to construct compact audio memory}
    \resizebox{0.9\linewidth}{!}{
    \begin{tabular}{cc}
    \hline \hline
     \makecell{Methods} & WER(\%)  \\ \hline 
     Baseline & 46.1 \\ \hdashline
     \textbf{AKVSR} (CPC \cite{oord2018representation})   &  42.7   \\
     \textbf{AKVSR} (Wav2vec2.0 \cite{baevski2020wav2vec})   & 42.4   \\
     \textbf{AKVSR} (HuBERT \cite{hsu2021hubert})   & \textbf{41.6}    \\
     \hline \hline
    \end{tabular}}
    \label{tab:3}
\end{table}

\begin{table}[]
  \renewcommand{\arraystretch}{1.4}
  \renewcommand{\tabcolsep}{8mm}
    \centering
    \caption{Impact of refining non-linguistic factors such as speakers and noise of audio knowledge when complementing visual modality with audio modality}
    \resizebox{1.0\linewidth}{!}{
    \begin{tabular}{cc}
    \hline \hline
     Method  & WER(\%) \\ \hline 
     AV-HuBERT \cite{shi2022learning} \textbf{+ KD\cite{hinton2015distilling}} & \textbf{43.6} \\
     AV-HuBERT \cite{shi2022learning} \textbf{+ Auxiliary task\cite{ma2022visual}} & \textbf{43.4} \\ \hdashline
     AV-HuBERT \cite{shi2022learning} \textbf{+ AKVSR}  & \textbf{41.6}    \\
     \hline \hline
    \end{tabular}}
    \label{tab:4}
\end{table}


\subsection{Effect of different pretrained audio models on constructing compact audio memory}

In the previous section, the proposed AKVSR injects linguistic information into the VSR model and shows promising results outperforming the current state-of-the-art VSR model. In this section, to assess the effectiveness of different large-scale pretrained audio models, we create compact audio memory using CPC, Wav2vec2.0, and HuBERT and apply the compact audio memory to the VSR model, respectively. The results are shown in Table \ref{tab:3}. While the baseline model (\ie without compact audio memory and ABM) achieves 46.1\% WER, our proposed method adopted by CPC, Wav2vec2.0, and HuBERT accomplishes a WER of 42.7\%, 42.4\%, and 41.6\%, respectively. The results show that the proposed compact audio memory can store the linguistic information of any large-scale pretrained audio model and improves the performance of the existing VSR methods by supplementing visual modality with the information. We utilize the HuBERT audio model in other experiments as it achieves the best performance. 

\subsection{Effect of refining non-linguistic factors when transferring the audio knowledge.}

In this section, we verify the effect of refining non-linguistic factors. To verify this, we experiment with two methods \cite{ma2022visual, hinton2015distilling} utilizing KD and compare them to the proposed method. The results are shown in Table IV. Following the method \cite{hinton2015distilling}, we use pretraind ASR  model as a teacher network and utilize the AV-HuBERT as a student network. From guiding the pretrained ASR model, we achieve a WER of 43.6\%. The other experiment is transferring the knowledge between intermediate layers. The effectiveness of this method called auxiliary task recently is more effective than \cite{hinton2015distilling} and verified in multiple languages VSR in \cite{ma2022visual}. Through this approach, we accomplish 43.4\% WER. At this time, our proposed method achieve a 41.6\% WER by using the same training data by refining the non-linguistic factors such as speakers and noise of audio knowledge and transferring via ABM.

\begin{table}[]
    \centering
    \renewcommand{\arraystretch}{1.2}
    \renewcommand{\tabcolsep}{8mm}
    \caption{VSR performances according to the number of cross-attention layers in ABM }
    \resizebox{0.8\linewidth}{!}{
    \begin{tabular}{ccc}
    \hline \hline
     \# Cross-attention layer  & WER(\%) \\ \hline 
     1 &  41.8   \\ 
     2 &   \textbf{41.6}   \\
     3 &  41.9   \\
     4 &  41.8   \\ 
     \hline \hline
    \end{tabular}}
    \label{tab:5}
\end{table}

\subsection{Effect of according to different number of layers in Audio Bridging Module}
The proposed ABM can be composed of different numbers of cross-attention layers. In order to evaluate the effect of the number of layers, we build 4 variants of ABM by differing the number of layers from 1 to 4 and perform VSR on LRS3. The ablation result is shown in Table \ref{tab:5}.
By using ABM consisting of a single cross-attention layer, we can achieve a significant performance improvement, 4.3\% WER from the baseline. This result confirms that utilizing compact audio knowledge through the proposed memory is effective in VSR by complementing insufficient visual information with audio information provided by a large-scale pretrained audio model. The best result is obtained when 2 cross-attention layers are utilized for ABM and we found that increasing the number of layers to more than 2 does not give more performance gain. Therefore, we employ 2 cross-attention layers for ABM in other experiments. Moreover, by comparing the number of parameters of ABM with the baseline model, adding one cross-attention layer increases 2.4M parameters, which is 1.5\% of that of the baseline model. Therefore, the best-performed model (\ie, 2 cross-attention layers) just requires only 3\% additional parameters of the baseline while improving 9.76\% relative performance from the baseline.

\begin{table}[]
  \renewcommand{\arraystretch}{1.2}
  \renewcommand{\tabcolsep}{9mm}
    \centering
    \caption{Improving performances of existing VSR methods by applying the proposed AKVSR}
    \resizebox{0.9\linewidth}{!}{
    \begin{tabular}{cc}
    \hline \hline
     Method  & WER(\%) \\ \hline 
     TM-seq2seq \cite{afouras2018deep}   & 59.9   \\ 
     TM-seq2seq \cite{afouras2018deep} \textbf{+ AKVSR} & \textbf{54.5} \\ \hdashline
     AV-HuBERT \cite{shi2022learning} & 46.1   \\ 
     AV-HuBERT \cite{shi2022learning} \textbf{+ AKVSR}  & \textbf{41.6}    \\
     \hline \hline
    \end{tabular}}
    \label{tab:6}
\end{table}

\subsection{Effect of AKVSR on different VSR methods}
We conducted experiments to demonstrate that our proposed method, which incorporates compact audio memory and ABM, can be applied to other VSR methods. To validate this, we apply the proposed method to another popular VSR method, TM-seq2seq \cite{afouras2018deep}. This model is originally designed to apply VSR in wild videos. We re-implemented the TM-seq2seq and trained it using the curriculum learning approach detailed in \cite{afouras2018deep}. The results are shown in \ref{tab:6}. We can achieve a WER of 54.1\% by applying the proposed AKVSR at TM-seq2seq. In addition, when we adopt our proposed method to the state-of-the-art AV-HuBERT VSR model. Our implementation resulted in a 4.5\% WER improvement. As a result, the proposed method injects the linguistic information of a large-scale pretrained audio model into visual modality at the feature level and can brings improvements to various VSR methods.

\begin{table}[]
    \renewcommand{\arraystretch}{1.2}
    \renewcommand{\tabcolsep}{5mm}
    \centering
    \caption{Impact of the number of clusters on the results of sentence prediction. "A" denotes the ASR performance of using only discrete units of the memory. "V" represents the VSR performance of the baseline model when it adapts compact audio memory using the ABM.}
    \resizebox{1.0\linewidth}{!}{
    \begin{tabular}{cccc}
    \hline \hline
    \multirow{2}{*}{Methods} & \multirow{2}{*}{\# of clusters} & \multicolumn{2}{c}{WER(\%)} \\
    \cline{3-4}
     &  & A  & V  \\ \hline 
    
     \textbf{AKVSR} (HuBERT \cite{hsu2021hubert}) & 200 & 9.4 & \textbf{41.6}    \\
     \textbf{AKVSR} (HuBERT \cite{hsu2021hubert}) & 500 & 8.1 & 42.0    \\
     \textbf{AKVSR} (HuBERT \cite{hsu2021hubert}) & 1000 & \textbf{6.2} & 41.8   \\
     \hline \hline
    \end{tabular}}
    \label{tab:7}
\end{table}

\subsection{Effect of varying the number of clusters}
We evaluate how storing linguistic information in compact audio memory is affected by varying the number of distinct clusters used to organize the information. The results are shown in Table VII. We conduct three experiments using 200, 500, and 1000 clusters referring to that utilizing audio feature clustering \cite{shi2022learning, lakhotia2021generative}. In these experiments, we employ HuBERT to extract and cluster the audio features. After storing the linguistic information in trainable compact audio memory via ASR task on the LRS3 dataset, we achieve a WER of 9.4\% a WER of 8.1\% WER, and a WER of 6.2\%, respectively. The results show that if the number of clusters increases, the WER of ASR decrease. However, when we apply the compact audio memories to the baseline models, we observe that the memory using 200 clusters obtains the most 41.6\% WER of performance. We analyze these results as that using the 200 discrete units in the compact audio memory is more appropriate to find the best-matched audio representation via ABM from the visual feature than employing other compact audio memories consisting of 500 and 1000 discrete units.

\subsection{Effect of varying dimension of compact audio memory}
We evaluate how the discrete representation dimension in compact audio memory affects the VSR performance of the proposed method. To this end, we conduct three experiments, varying the dimensions of the compact audio memory to 512, 768, and 1024, when constructing the memory to store linguistic information from audio. Then, these memories are applied to the baseline VSR model. The results are shown in Table \mbox{\ref{tab:8}}. When the compact audio memory of 768 dimensions is used, we obtain the best performance, a WER of 41.5\%. The other settings using 512 and 1024 dimensions achieve 42.0\% and 41.7\% WERs, respectively. Please note that all experiments improve the performance of the baseline VSR model achieving a WER of 46.1\%.

\subsection{Effect of varying training dataset when constructing the compact audio memory}
We verify how the varying datasets for training the compact audio memory affect the performance of the baseline. For this, LRS2, LRS3, and merging LRS2 with LRS3 are used for building compact audio memories. The results are shown in Table \mbox{\ref{tab:9}}. In these settings, we obtain 44.7\%, 43.3\%, and 41.5\% WERs, respectively. Based on these results, the compact audio memory trained on merging LRS2 with LRS3 datasets is applied to the VSR model in other experiments.

\begin{table}[]
    \renewcommand{\arraystretch}{0.99}
    \renewcommand{\tabcolsep}{20mm}
    \centering
    \caption{Ablation Study on the Proposed Method Performance with Varying Dimension of Compact Audio Memory.}
    \resizebox{1.0\linewidth}{!}{
    \begin{tabular}{cc}
    \hline \hline
    Dimension & WER(\%) \\ \hline  
     512 & 41.7    \\
     768 & \textbf{41.5}    \\
     1024 & 42.0  \\
     \hline \hline
    \end{tabular}}
    \label{tab:8}
\end{table}

\begin{table}[]
    \renewcommand{\arraystretch}{1.0}
    \renewcommand{\tabcolsep}{8mm}
    \centering
    \caption{Ablation Studies on the Effects of Varying Training Datasets in the Construction of Compact Audio Memory}
    \resizebox{1.0\linewidth}{!}{
    \begin{tabular}{ccc}
    \hline \hline
    Training Datasets & Duration (hrs) & WER(\%) \\ \hline  
     LRS2 & 223 & 44.7    \\
     LRS3 & 433 & 43.3    \\
     LRS2, LRS3 & 656 & \textbf{41.5}  \\
     \hline \hline
    \end{tabular}}
    \label{tab:9}
\end{table}

\section{Conclusion}
This paper has presented the Audio Knowledge empowered Visual Speech Recognition (AKVSR) framework to enhance the complementary effect of audio knowledge for visual modality, by removing the non-linguistic factors. This proposed framework consists of three components: (1) Unlike previous methods, the proposed approach utilizes a large-scale pretrained audio model to acquire audio knowledge, and the non-linguistic factors, such as speaker and noise, are then removed through vector quantization from the audio knowledge. (2) The resulting audio knowledge, focusing on linguistic information, has been stored in a compact audio memory to allow for the utilization of audio information. (3) The Audio Bridging Module has been devised to match the best audio knowledge with the visual features in the compact audio memory to complement the insufficient visual information and inject linguistic information into the VSR model. We have demonstrated the effectiveness of the proposed method by outperforming the state-of-the-art VSR models on the LRS3 dataset.

\section{Discussion}

Although the proposed AKVSR can improve the VSR systems using the audio knowledge of large-scale pretrained audio models, there is some limitation and future research: (1) In this paper, the compact audio memory needs to be constructed before training the VSR model. It requires additional time consumption. In future research, we are going to further investigate a time-efficient training method. (3) Compared with the ASR systems, the performance of the VSR system (including the proposed method) is still inferior. To bridge the thin gap between VSR and ASR, in future research, how to utilize the audio model needs to be continually investigated.

\section{Acknowledgment}
This work was supported in part by the IITP grant funded by the Korea government (MSIT) (No.2020-0-00004, Development of Previsional Intelligence based on Long-Term Visual Memory Network), in part by the National Research Foundation of Korea (NRF) grant funded by the Korea government (MSIT) (No.NRF-2022R1A2C2005529), and in part by the BK21 FOUR(Connected AI Education \& Research Program for Industry and Society Innovation, KAIST EE, No. 4120200113769).

\bibliographystyle{IEEEtran}

\bibliography{egbib}

\begin{IEEEbiography}[{\includegraphics[width=1in,height=1.25in,clip]{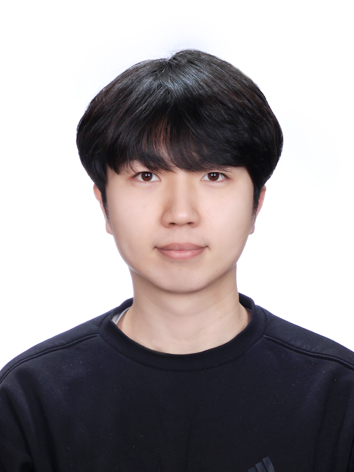}}]{Jeong Hun Yeo}
received the B.S. and M.S. degrees in electrical \& electronic engineering from Korea Advanced Institute of Science and Technology (KAIST), Daejeon, South Korea, in 2020 and 2022, respectively. He is currently pursuing the Ph.D. degree in electrical engineering at KAIST, Daejeon, South Korea. His research interests include deep learning, image/video analysis, visual speech recognition, and multi-modal learning.
\end{IEEEbiography}

\begin{IEEEbiography}[{\includegraphics[width=1in,height=1.25in,clip]{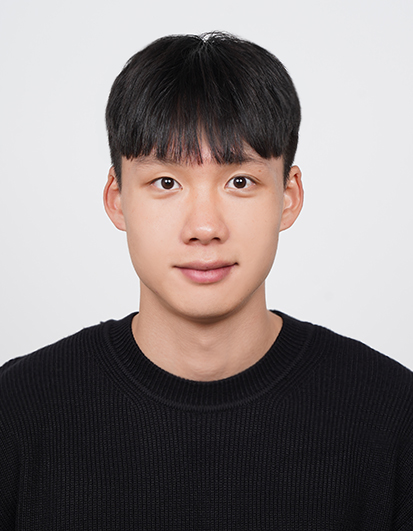}}]{Minsu Kim}
received the B.S. degree in electrical \& electronic engineering from Yonsei University, Seoul, South Korea, in 2019. He is currently pursuing the Ph.D. degree in electrical engineering at Korea Advanced Institute of Science and Technology (KAIST), Daejeon, South Korea. His research interests include deep learning, image/video analysis, audio-visual speech recognition, and multi-modal analysis.
\end{IEEEbiography}

\begin{IEEEbiography}[{\includegraphics[width=1in,height=1.25in,clip]{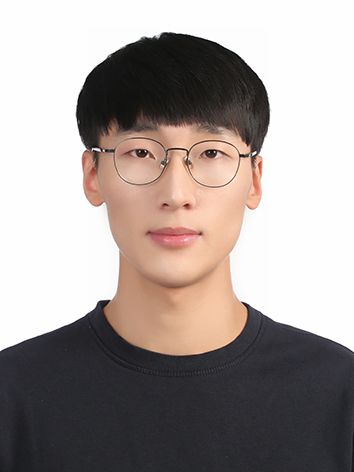}}]{Jeongsoo Choi}
received the B.S. degree in electrical \& electronic engineering from Korea Advanced Institute of Science and Technology (KAIST), Daejeon, South Korea, in 2020. He is currently pursuing the Ph.D. degree in electrical engineering at KAIST, Daejeon, South Korea. His research interests include deep learning, image/video analysis, speech synthesis, and multi-modal analysis.
\end{IEEEbiography}

\begin{IEEEbiography}[{\includegraphics[width=1in,height=1.25in,clip]{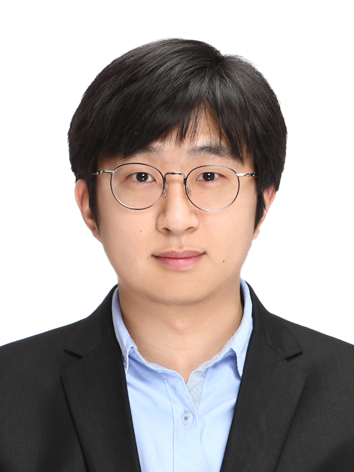}}]{Dae Hoe Kim}
received the B.S. degree from Hanyang University, Seoul, Korea, in 2010 and the M.S. and Ph.D. degrees from the Korea Advanced Institute of Science and Technology (KAIST), Daejeon, South Korea, in 2012 and 2017, respectively. He was a Senior Researcher with Agency for Defense Development, Daejeon, South Korea. Since 2019,  he has been working as a Senior Researcher at the Superintelligence Creative Research Laboratory of the Electronics and Telecommunications Research Institute, Daejeon. His research interests include visual recognition, video action localization, machine learning, and computer vision.
\end{IEEEbiography}

\begin{IEEEbiography}[{\includegraphics[width=1in,height=1.25in,clip]{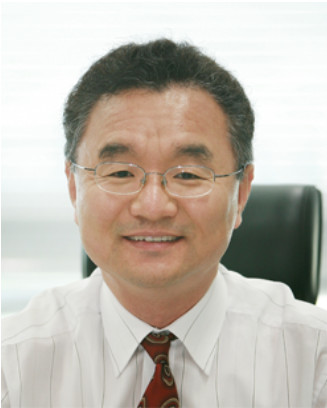}}]{Yong Man Ro}
(Senior Member, IEEE) received the B.S. degree from Yonsei University, Seoul, South Korea, and the M.S. and Ph.D. degrees from the Korea Advanced Institute of Science and Technology (KAIST), Daejeon, South Korea. He was a Researcher at Columbia University, a Visiting Researcher at the University of California at Irvine, Irvine, CA, USA, and a Research Fellow of the University of California at Berkeley, Berkeley, CA, USA. He was a Visiting Professor with the Department of Electrical and Computer Engineering, University of Toronto, Canada. He is currently a Professor with the Department of Electrical Engineering and the Director of the Center for Applied Research in Artificial Intelligence (CARAI), KAIST. Among the years, he has been conducting research in a wide spectrum of image and video systems research topics. Among those topics, his interests include image processing, computer vision, visual recognition, multimodal learning, video representation/compression, and object detection. He received the Young Investigator Finalist Award of ISMRM, in 1992, and the Year’s Scientist Award (Korea), in 2003. He served as an Associate Editor for IEEE Signal Processing Letters. He currently serves as an Associate Editor for IEEE Transactions on Circuits and Systems for Video Technology. He served as a TPC in many international conferences, including the program chair and organized special sessions.
\end{IEEEbiography}

\vfill

\end{document}